\documentclass[11pt,a4paper]{article} 
\usepackage[left=2cm,right=2cm,top=2cm,bottom=2cm]{geometry}
\usepackage{amsmath,amsthm,amstext,amsfonts,amssymb,mathrsfs}
\usepackage{graphicx,subfigure,algorithm,algorithmic}
\usepackage{tikz}

\newtheorem{definition}{Definition}
\newtheorem{theorem}{Theorem}
\newtheorem{proposition}{Proposition}

\newtheorem{assumption}{Assumption}

\theoremstyle{remark}
\newtheorem{remark}{Remark}

\title{Approachability in Stackelberg Stochastic  Games with Vector Costs}
\author{Dileep Kalathil, Vivek S. Borkar and Rahul Jain}
\date{}

\begin{document}

\maketitle

\begin{abstract}
The notion of approachability was introduced by Blackwell \cite{blackwell1956} in the context of vector-valued repeated games. The famous `Blackwell's approachability theorem' prescribes a strategy for approachability, i.e., for `steering' the average vector cost of a given agent towards a given target set, irrespective of the strategies of the other agents. In this paper, motivated by the multi-objective optimization/decision making problems in dynamically changing environments, we address the approachability problem in  Stackelberg stochastic games with vector-valued cost functions. We make two main contributions. Firstly, we give a simple and  computationally  tractable  strategy for approachability for Stackelberg stochastic games along the lines of Blackwell's. Secondly, we give a reinforcement learning  algorithm for learning the approachable strategy when the transition kernel  is unknown. We also recover  as a by-product Blackwell's necessary and sufficient conditions for approachability  for convex sets in this set up and thus a complete characterization. We  give sufficient conditions for non-convex sets.
%
%
\end{abstract}

\section{Introduction}
\label{sec:Intro}

Classical game theory, Markov Decision Processes (MDPs) and stochastic games  typically deal only with scalar performance criteria: corresponding to each state and action of the agents, each agent incurs a \emph{scalar cost}. The standard problem is to compute or \emph{learn}  a policy for each agent which will minimize her  scalar performance  objective conditioned on the policies of all other agents and the dynamics of the underlying system. Computing or learning  equilibrium policies in a standard normal form strategic game or stochastic game, computing or learning the optimal policy of average or discounted cost MDPs, are the typical examples. However,   many interesting problems often fall outside this `scalar performance criteria' class. For example, consider the problem faced by automobile manufacturers. They want to minimize their cost but also worry about reliability, perceived quality and customer satisfaction - all of which are quantified in some way and are required to be greater than some prescribed values. Bandwidth allocation in a wireless network is another instance which involves throughput maximization while also providing certain delay guarantees.

Multi-objective optimization is a well studied area \cite{steuer1989multiple} though most methods focus on achieving  a Pareto-optimal solution. In the context of decision making under dynamically changing environment, these problems have  been studied extensively under the rubric `constrained MDPs' \cite{altman1999constrained}. A reinforcement learning algorithm for constrained MDPs was developed in \cite{borkar2005actor}. Articles \cite{yu2009markov} and \cite{even2009online} consider  MDPs with arbitrary reward process. Their setting is in the framework of regret minimization which is different from our approach.  Approachability for repeated games has been analyzed extensively in the literature; we refer the reader to the survey paper \cite{perchet2014approachability} and the references   therein.

Approachability in a stochastic game framework has also been addressed before.  Article \cite{shimkin1993Blackwell} addressed this   problem   where the approachability from a given initial state was studied under some recurrence assumptions for countable state space  controlled Markov chains. Their scheme depends on updating strategies when the system returns to a fixed  state $s_{0}$. This scheme was proposed because there appeared to be a need to keep the policy  fixed for some duration in order to `exploit' that policy before one `explores' again. However, there are many computational difficulties associated with this approach, in particular slow speed if the returns to the prescribed  fixed state are infrequent,  a common situation in large problems. Furthermore, this approach has the undesirable effect of increasing the variance of the cost for the agents if the recurrence times are large, again a typical scenario in very large systems. For this reason, the scheme will have slow convergence and high fluctuations. In \cite{milman2006Blackwell}, an alternative scheme is proposed that requires less restrictive assumptions. The basic idea is the `increasing time window' method:  keep the policy constant for a length of a time window whose duration increases gradually. In each window, the policy used is the equilibrium policy of an $N_{i}$-stage stochastic game where $N_{i}$ is the length of the $i$th window. The computation of this policy is thus clearly non-trivial.  In turn, \cite{Sameer} presents yet another scheme which has similar drawbacks. These schemes also do not address another important aspect of this problem: a learning algorithm  for approachability when the transition kernel corresponding to the underlying Markov dynamics is unknown. In \cite{mannor2004geometric}, one does have a learning algorithm that builds on the work in \cite{shimkin1993Blackwell}. The key idea is to run  $J$ learning algorithms in parallel, each one corresponds to a different steering direction. If $J$ is sufficiently large,  $\epsilon$-approachability can be guaranteed. This scheme, however, is potentially computationally impractical in some scenarios.

The key difference in the present work is our adaptation of the Stackelberg framework which, by setting a much less ambitious goal,  sidesteps the main difficulties faced in the above works and facilitates a result closer in spirit to the original Blackwell formulation. In Stackelberg stochastic games, one agent (leader) takes an action first and the second agent (follower) takes the action next after observing the action of the leader. However, the evolution of the state depends on the action of both agents. A precise definition is given in the next section. For Stackelberg stochastic games, we recover the necessary and sufficient conditions for approachability of convex sets and sufficient conditions for non-convex sets. As expected, these conditions are similar to prior work on repeated and Markov  cases, but our proof is different as we rely on ideas from stochastic approximation theory for constructing an approachability strategy. Stochastic approximation as a tool for analyzing approachability was also used in \cite{BenHofSor1},  \cite{BenHofSor2}  where approachability for repeated games has been analyzed extensively.

The Markov case considered here and in  \cite{shimkin1993Blackwell}  \cite{milman2006Blackwell} \cite{Sameer} is significantly more complicated because the dynamics interferes with the simple strategy of Blackwell. The works  \cite{shimkin1993Blackwell}  \cite{milman2006Blackwell} \cite{Sameer} consider general non-cooperative stochastic games and establish analogous conditions for approachability as well as optimal strategies. As mentioned above, they consider strategies which include a `waiting' component, i.e., the updates are done only along a sub-sample, not every time. They differ in the manner the waiting is effected. This `waiting' facilitates learning by allowing pure exploration. As they consider the general non-cooperative case,   our approachability results are weaker insofar as they are restricted to Stackelberg games.  This way we impose an order in which the agents operate and allow the follower to know the leader's move, thus modifying the information structure. As will become apparent later, this essentially makes it a combination of a zero sum game and a Markov decision process, both vastly more amenable to analysis than the general non-cooperative case. This significantly simplifies the analysis. The primary gain is that one now has a strategy \textit{sans} any waiting, because updates can be performed at each step. This takes away the potential degradation of convergence speed due to waiting. The flip side is that this is a much more restrictive model.

Using multiple time scale stochastic approximation theory, we are also able to give a `learning' scheme for Stackelberg stochastic games when the transition kernel is unknown. It appears difficult to derive a learning scheme from approachability strategies for stochastic games from any of the prior works (except \cite{mannor2004geometric}). Thus, our main contributions are: (i) a new strategy  for approachability for Stackelberg stochastic games recovering known characterizations in other paradigms for this case, and (ii) stochastic approximation-based learning schemes for approachability when the transition kernel is unknown.

We describe the problem in detail in the next section. A sufficient/necessary conditions and a strategy for approachability of a convex set are specified in Section 3.  Section 4 extends these to non-convex targets. Section 5 describes a learning algorithm for approachability. Section 6 concludes the paper with a brief remark about future directions.

\section{Preliminaries}

Consider a  stochastic game with two agents,  finite state space $\mathcal{S}$, finite action space $\mathcal{A} = \mathcal{A}^{1} \times \mathcal{A}^{2}$ where $\mathcal{A}^{i}$ is the action space of agent $i, i=1,2$. We shall denote by $\{s_n, n \geq 0\}$ the $\mathcal{S}$-valued state process. An element $a=(a^{1}, a^{2}) \in \mathcal{A}$ is called an action vector. Let $p(\cdot, \cdot, \cdot)$ be the transition kernel that governs the state evolution, $p(s, a, \cdot) \in   \mathcal{P}(\mathcal{S})$ for all $s \in  \mathcal{S}, a \in \mathcal{A}$ where $\mathcal{P}(\mathcal{S})$ is the set of probability distributions over the set $\mathcal{S}$.
Let $\underline{c} : \mathcal{S} \times \mathcal{A} \rightarrow \mathbb{R}^{K}$ be the \emph{vector-cost} function for agent $1$, and  for a given state $s \in \mathcal{S} $ and an action vector $a \in \mathcal{A}$, $\underline{c}(s,a)=[c_{1}(s,a), \ldots, c_{K}(s,a)]^{\dagger}$ where $c_{j} : \mathcal{S} \times \mathcal{A} \rightarrow \mathbb{R}$ for $1 \leq j \leq K$. We assume that the  cost function is  bounded  and without loss of generality assume that $|c_{j}(s,a)| \leq 1, 1 \leq j \leq K, 1 \leq i \leq 2, \forall s \in \mathcal{S}, \forall a \in \mathcal{A}$.  The average vector-cost incurred for agent $1$ till time $n$ is denoted by $x_{n} = \frac{1}{n}\sum^{n}_{m=1}\underline{c}(s_{m},a_{m})$, where $\{a_n = (a^1_n, a^2_n)\}$ is the control sequence being used. Note that   this  stochastic game reduces to a multi-objective MDP when $\mathcal{A}^{2}$ is a singleton. Throughout, $\mathcal{P}(X)$ for a Polish space $X$ will denote the Polish space of probability measures on $X$ with the Prohorov topology.

\begin{remark}
In a more general setting, the action space $\mathcal{A}^{i}$ of each agent $i$ can depend on the current state $s \in \mathcal{S}$ and is denoted by $\mathcal{A}^{i}(s)$. However, it is not difficult to show that all results in this work can be extended to the case where the action space is state dependent at cost of some additional notational complexity.  This is  straightforward from the corresponding results in MDPs and zero-sum stochastic games \cite{filar1996competitive}. We restrict to the former model for simplicity of notation.
\end{remark}

In many situations, the transition probabilities are either unknown or too complex to afford explicit analytic handling of the dynamic programming equations, but a simulation device or actual experimentation can provide input-output data (e.g., network simulators, robotics). In such cases a popular approach to approximate dynamic programming is to use the so called reinforcement learning algorithms, which essentially are stochastic approximation schemes to solve the dynamic programming equations approximately. Developing such a learning algorithm for a general non-cooperative stochastic game is known to be difficult except in very special cases such as zero-sum \cite{patek1997}.  Here  we consider a natural relaxation of the problem as `game against nature', and formalize it as  a \emph{Stackelberg stochastic game}. In a Stackelberg stochastic game, at each time step $n$, agent $1$ takes an action $a^{1}_{n}$  first. Player $2$ (adversary)  observes this action and then  selects her action $a^{2}_{n}$.

Recall that a behavioral strategy  for agent $i$ is a sequence of functions which successively map the available `history' at any given time $m$ to  a probability distribution on $\mathcal{A}_i$ which gives the conditional probability with which the  action $a^i_{m} \in \mathcal{A}_i$ is chosen. In a Stackelberg stochastic game,  agent $1$'s behavioral strategy  $\sigma^{1} = [\sigma^{1}_0, \sigma^{1}_1, \cdots]$ is a sequence of maps $\sigma^{1}_m: \mathcal{S}^{m+1}\times\mathcal{A}_1^m \mapsto \mathcal{P}(\mathcal{A}_1)$  and  agent $2$'s (adversary) behavioral strategy  $\sigma^{2} = [\sigma^{2}_0, \sigma^{2}_1, \cdots]$ is a sequence of maps $\sigma^{2}_m: \mathcal{S}^{m+1}\times\mathcal{A}^m \times \mathcal{A}_{1} \mapsto \mathcal{P}(\mathcal{A}_2)$, because she also observes agent 1's actions. When these conditional probabilities  depend only on the current state for agent $1$ and on the current state and agent $1$'s current action for agent $2$, we shall call them stationary  strategies and identify them with maps $\pi^1: \mathcal{S} \rightarrow \mathcal{P}(\mathcal{A}_1)$ and $\pi^2: \mathcal{S}\times\mathcal{A}^1 \rightarrow \mathcal{P}(\mathcal{A}^2)$, resp.  Let $\Sigma^{i}$  be the set of  behavioral strategies of agent $i, i=1,2$. A pair of strategies $(\sigma^{1}, \sigma^{2}) \in \Sigma^{1} \times \Sigma^{2} $ together with an initial state $s_0$ induces a probability distribution $\mu_{s_0}(\sigma^{1}, \sigma^{2})$ on the sequence of vectors $\{x_{n}\}$.

Our objective is to specify  the conditions under which a given closed set is \emph{approachable}, propose a  strategy for approachability and prescribe a learning algorithm  for approachability. The notion of an approachable set for stochastic games is made precise in the following definition.

\begin{definition}[Approachable Set]
 A closed  set $D$ is approachable for agent $1$ if there exists a behavioral strategy $\sigma^{1}$ for agent $1$ such that $\|x_{n} - D\| \rightarrow 0$ as $ n \rightarrow \infty$, $\mu_{s_0}(\sigma^{1}, \sigma^{2})$-almost surely for all $\sigma^{2} \in \Sigma^{2}$, and for all initial states
  $s_0 \in \mathcal{S}$.
\end{definition}

Let $\Psi$ denote the set of ergodic occupation measures over the set $\mathcal{S}\times\mathcal{A}$. That is, any $\psi \in \Psi$ can be decomposed as $\psi(s, a^{1}, a^{2}) = \eta^{\psi}(s) \pi^{1}(a^1|s) \pi^{2}(a^2|s, a^{1})$ where:
  \begin{itemize}
  \item $\pi^{1}(\cdot|s) \in \mathcal{P}(\mathcal{A}^{1})$, $\pi^{2}(\cdot|s, a^{1}) \in \mathcal{P}(\mathcal{A}^{2})$, and,

   \item $\eta^{\psi}$ is an invariant probability measure over $\mathcal{S}$ induced by the stationary policy pair $(\pi^{1}, \pi^{2})$ and the controlled transition kernel $p(\cdot, \cdot, \cdot)$.
\end{itemize}

For any $\psi \in \Psi$,  define the \emph{ergodic vector-cost} function for agent $i$ as
\begin{equation}
\label{eq:ergodiccost-game}
\underline{c}(\psi): \Psi \rightarrow \mathbb{R}^{K} = \left[\sum_{s,a^{1},a^{2}} \psi(s,a^{1},a^{2}) c_{1}(s,a^{1},a^{2}), \ldots, \sum_{s,a^{1},a^{2}} \psi(s,a^{1},a^{2}) c_{K}(s,a^{1},a^{2}) \right]^{\dagger}
\end{equation}
Note that $\underline{c}(\psi)$  is the expected average vector-cost for agent $i$ if both agents play a pair of stationary strategies  $(\pi^{1}, \pi^{2})$ such that  $\psi$ is the ergodic occupation measure induced by $(\pi^{1}, \pi^{2})$.

\noindent \textbf{Remark:} More generally, we may consider general behavioral strategies for either of both agents. However, it follows from the results of \cite{borkar1991topics}, Chapter VI (see the proof of Theorem 1.1, pp.\ 58-59) that under an arbitrary behavioral strategy pair, every limit point of $\{x_n\}$ as $n\uparrow\infty$ will be of the form $\underline{c}(\psi)$ for some ergodic occupation measure $\psi$. Thus there is no loss of attainable payoffs by restricting a priori to stationary Markov policies $(\pi^1, \pi^2)$ and moreover, the latter have the appeal of lower memory requirements.

We make the following assumption.
\begin{assumption}
\label{assumption1-sg}
The Markov chain induced by any pair of  stationary strategies $(\pi^{1}, \pi^{2})$ by the agents  is irreducible.
\end{assumption}
This is a standard assumption made in reinforcement learning. The basic assumption in \cite{shimkin1993Blackwell} is that there exists a state $s_{0}$ such that the mean first-passage time from any arbitrary state to $s_{0}$ is finite for all stationary strategies of the agents. This is the so called `uni-chain' case. The  consequences of Assumption 1 that are of relevance here are (\cite{puterman2014markov}, Chapter 8):
 \begin{enumerate}
 \item the existence of a single communicating class under any stationary strategy,
 \item accessibility (i.e., a.s. finite hitting time) for the communicating class from any state under any stationary strategy,
 \item a unique stationary distribution under any stationary strategy (implying in particular a unique ergodic occupation measure),
 \item an optimal stationary strategy exists that is optimal for any initial condition.
 \end{enumerate}
 In fact, under Assumption 1, the whole state space is a communicating class, whereby the above consequences are trivial. But they also hold under the seemingly weaker uni-chain hypothesis (\cite{puterman2014markov}, Chapter 8).
 For the sake of completeness, we state the uni-chain assumption below:

\begin{assumption}[Uni-chain MDP] 
\label{ass:unichain}
An MDP is uni-chain if the transition matrix corresponding to every deterministic stationary policy is uni-chain, that is, it consists of a single recurrent class plus a possibly empty set of transient states. 
\end{assumption}
Our core results for approachability will hold for the uni-chain case as well. The key fact that ensures this is that under any stationary strategy the invariant distribution and therefore the ergodic occupation measure is uniquely specified.
Where we need the full strength of Assumption 1 is when we develop the learning algorithm, which requires sufficient `exploration' for learning, as will become clear later. For the same reason, we also confine ourselves to a finite state space case, though the results concerning approachability alone will go through for countable state space as in, e.g., \cite{Sameer}, under additional conditions that either ensure blanket stability (such as a suitable stochastic Liapunov condition) or a suitable condition on costs that penalizes instability - see, e.g., Chapters VI, VII of \cite{borkar1991topics}. In \cite{mannor2003empirical} too   a similar but weaker assumption is used. Our Assumption 1 is thus stronger than that in \cite{shimkin1993Blackwell}  \cite{mannor2003empirical}.

\section{Conditions for Approachability of Convex Sets}
\label{sec:cond-suff-necc}

In this section we give the necessary and sufficient conditions for approachability of convex sets. We note that these  conditions have been given in previous work as well \cite{shimkin1993Blackwell} \cite{milman2006Blackwell}. Our main contribution here is to give a new approachable strategy using stochastic approximation techniques that permits continuous updates, albeit only for the special case of Stackelberg games. The main motivation for providing a new approachable strategy, apart from the computational simplicity of the proposed strategy, is that it is amenable to a learning scheme. We indeed use this strategy  to propose a learning scheme for approachability in Section \ref{sec:learning}. We assume that Assumption \ref{ass:unichain} is true throughout this section.

For now we restrict our attention  to a convex set $D$. Extension to a non-convex set is given in Section \ref{sec:non-convex-sg}.
We begin  by defining a few quantities.  For any  $x \in \mathbb{R}^{K} \setminus D$, let $P_{D}(x)$ denote the (unique) projection of $x$ onto $D$. Let $\lambda(x)= \left(x - P_{D}(x)\right)/\|x - P_{D}(x)\| $. For every $x \in \mathbb{R}^{K} \setminus D$, we define a scalar-valued Stackelberg stochastic  game with the stage cost $\tilde{c}(s, a;x)=\langle  \underline{c}(s, a), \lambda(x) \rangle$ where $\langle \cdot, \cdot \rangle$ denotes the inner product between two vectors. With respect to this scalar-valued Stackelberg stochastic  game, we define the following quantities:
\begin{align}
\label{eq:minmaxcost-sg}
{c}^{*}(x) &=  \min_{\pi^{1}} \max_{\pi^{2}} \{\langle \underline{c}(\psi), \lambda(x) \rangle : \psi(s,a^{1},a^{2}) = \eta^{\psi}(s) \pi^{2}(a^{2}|s,a^{1}) \pi^{1}(a^{1}|s)  \} \\
\label{eq:Gamma}
\Gamma_{\pi^1}(x) &= \arg \max_{\pi^2} ~~ \{\langle \underline{c}(\psi), \lambda(x) \rangle : \psi(s,a^{1},a^{2}) = \eta^{\psi}(s) \pi^{2}(a^{2}|s,a^{1}) \pi^{1}(a^{1}|s)  \} \\
\label{eq:Pi1(x)-sg}
\Pi^{1}(x) &= \arg \min_{\pi^{1}} ~~ \max_{\pi^{2}} \{\langle \underline{c}(\psi), \lambda(x) \rangle : \psi(s,a^{1},a^{2}) = \eta^{\psi}(s) \pi^{2}(a^{2}|s,a^{1}) \pi^{1}(a^{1}|s)  \} \\
\label{eq:Psi(x)-sg}
\Psi(x) &= \{\psi \in \Psi: \psi(s,a^{1},a^{2}) = \eta^{\psi}(s) \pi^{2}(a^{2}|s,a^{1}) \pi^{1}(a^{1}|s), \pi^{1} \in \Pi^{1}(x), \pi^2 \in \Gamma_{\pi^1}(x) \} \\
\label{eq:tildaPsi(x)-sg}
\tilde{\Psi}(x) &= \{\psi \in \Psi: \langle \underline{c}(\psi), \lambda(x) \rangle \leq c^{*}(x) \}
\end{align}

Given an $x \in \mathbb{R}^{K} \setminus D$,   $\Gamma_{\pi^1}(x)$ can be interpreted as the set of stationary strategies of a worst case adversary (agent 2) and  $\Pi^{1}(x)$ can be thought as the set of best stationary strategies of agent 1 against this adversary. $\Psi(x)$ is  the ergodic occupation measures corresponding to these strategies.

Our main result of this section is the following theorem. The main content is part $(ii)$ which specifies a strategy for approachability. Part $(i)$  and the necessary conditions that follow (Proposition \ref{prop:necess-sg}) are known results \cite{shimkin1993Blackwell}. They are  included here  also because the proofs have a bearing on what follows. Note that together, these provide conditions for approachability that are both necessary and sufficient.

\begin{theorem}
\label{thm:bat-avggame}
(i) (Sufficient Condition) A closed convex set $D$ is approachable from all initial states in a Stackelberg  stochastic game satisfying Assumption \ref{assumption1-sg}  if for every $x \in \mathbb{R}^{K} \setminus D$, there exists a (possibly non-unique) occupation measure $\psi$, $\psi(s,a^{1},a^{2})=\eta^{\psi}(s) \pi^{2}(a^{2}|s,a^{1}) \pi^{1}(a^{1}|s)$, such that  $\underline{c}(\psi) = c^*(x)$ and  $\langle \underline{c}(\psi) - P_{D}(x) , \lambda(x)\rangle \leq 0$.

(ii) (Strategy for Approachability) A strategy for approachability is:  At every time instant $n$, select action $a^{1}_{n}$ according to a policy $\pi^{1}( \cdot | x_{n}) \in \Pi^{1}(x_{n})$.
\end{theorem}

The proof is based on stochastic approximation theory \cite{Bo08}. The average vector-cost till time step $n+1$ can be written as
\begin{align}
\label{eq:saeqn1-sg}
x_{n+1} &= \frac{1}{n+1}\sum^{n+1}_{m=0}\underline{c}(s_{m},a_{m}) \\
&= x_{n} + \gamma(n) (\underline{c}(s_{n+1},a_{n+1})-x_{n})  \nonumber \\
&= x_{n} + \gamma(n) \left(\left(\underline{c}(s_{n+1},\pi^1(\cdot|x_{n}), a^{2}_{n+1})-x_{n}\right) + M_{n+1}\right),
\end{align}
where
\begin{eqnarray}
\gamma(n) &=& 1/(n+1), \nonumber\\
\underline{c}(s_{n+1},\pi^{1}(\cdot|x_{n}), a^{2}_{n+1}) &=& \sum_{a^{1} \in \mathcal{A}^{1}} \pi^{1}(a^{1} | x_n)  \underline{c}(s_{n+1},a^{1}, a^{2}_{n+1}), \nonumber \\
  \mbox{and}  \ \ M_{n+1} &=& \left(\underline{c}(s_{n+1},a^{1}_{n+1}, a^{2}_{n+1}) -  \underline{c}(s_{n+1},\pi(\cdot|x_{n}), a^{2}_{n+1})\right). \label{mg}
    \end{eqnarray}
    The key idea in  the analysis of \eqref{eq:saeqn1-sg} is to show  that it asymptotically tracks the   differential inclusion dynamics given by the differential inclusion
\begin{equation}
\label{eq:diffincleq1-sg}
\dot{x}(t) \in w(x(t)) - x(t),
\end{equation}
where
\[w(x) :=  \overline{co}\left(\{\underline{c}(\tilde{\psi}): \tilde{\psi} \in \tilde{\Psi}(x)\}\right) \]
and $\overline{co}(\cdot)$ denotes the closed convex hull of a set. Then, by showing that the dynamics given by \eqref{eq:diffincleq1-sg} converges to the set $D$,  we will conclude that the sequence $\{x_{n}\}$ also converges to the same set a.s.

\noindent We start with the following proposition.
\begin{proposition}
\label{thm:convexityetc}
The sets $\Psi(x)$ and $\tilde{\Psi}(x)$ are compact, non-empty and the set-valued maps $\chi: x \mapsto \Psi(x), \tilde{\chi}: x \mapsto \tilde{\Psi}(x)$ are upper semi-continuous.
\end{proposition}

\proof
Consider the Stackelberg stochastic game wherein at each time instant $n$, the state is $s_{n} \in \mathcal{S}$ and agent $1$ moves first and generates a randomized action $a^{1}_{n} \in \mathcal{A}^{1}$ according to conditional law $\pi^1( \cdot |s_n)$, followed by agent $2$ who observes $a^1_n$ and chooses a randomized action according to the conditional law $\pi^2( \cdot | s_n, a^1_n)$. Thus agent $2$ faces an average reward MDP with state process $(s_n, a^1_n), n \geq 0$, and transition probability
$\tilde{p}((\tilde{s}, \tilde{a}^{1}) | (s, a^{1}), a^{2}) := p(s, (a^{1}, a^{2}), \tilde{s})\pi^1(\tilde{a}^{1} | \tilde{s})$. For a given  $x \in \mathbb{R}^{K} \setminus D$, the optimal adversarial policy $\pi^{2}$ by agent $2$ for the scalar-valued MDP with cost function $\langle \underline{c}(s, a), \lambda(x) \rangle$ can be characterized as the  maximizer in the associated dynamic programming equation
\begin{equation}
V((s,a^{1})) = \max_{\pi^2}\left(\sum_{(\tilde{s}, \tilde{a}^{1})}  \sum_{a^{2}} \pi^2(a^{2} | s, a^{1})\tilde{p}((\tilde{s}, \tilde{a}^{1}) | (s, a^{1}), a^{2}) \left(\langle \underline{c}(s, a^{1}, a^{2}), \lambda(x)\rangle + V((\tilde{s}, \tilde{a}^{1})) \right) \right) - \beta. \label{firstDP}
\end{equation}
This has a solution $(V, \beta)$ where $\beta$ is unique and $V$ is unique upto an additive scalar and can be rendered unique by arbitrarily setting $V((s_{0}, a^{1}_{0}))=0$ for some $s_0, a^1_0$. Then $\Gamma_{\pi^1}(x)$ defined above is simply the set of maximizers on the R.H.S. Being the set of maximizers of an affine function on a convex compact set, it is a non-empty convex compact set. We see below that it is an upper semicontinuous set valued map as a function of $x$.\\

Now consider the MDP with state process $\{s_n\}$ and stationary randomized policies $\pi((a^{1}, a^{2}) | s)$  $ := \pi^{1}(a^{1} | s)\pi^{2}(a^{2} | s, a^{1})$ where $\pi^{1}(  \cdot | s) \in \mathcal{P}(\mathcal{A}^{1})$ and $\pi^2( \cdot | s, a^{1}) \in \Gamma_{\pi^1}(x)$. That is, the action space for the problem is $\Phi := \cup_{\pi^1 \in \mathcal{P}(\mathcal{A}^{1})}\left(\{\pi^1\}\times\Gamma_{\pi^1}(x)\right)$ with the relative topology inherited from $\mathcal{P}(\mathcal{A}^{1}\times\mathcal{A}^{2})$. Being a closed subset of a compact space, it is compact. By our definition of $\Gamma_{\pi^1}(x)$, the optimal policy for this MDP corresponds to the min-max policy for the original problem, i.e., belongs to $\Pi^1(x)$. This in turn is given by the minimizers on the right hand side of the dynamic programming equation
\begin{equation}
\hat{V}(s) = \min_{\pi \in \Phi}\left(\sum_{\tilde{s}} \sum_{a^{1}, a^{2}}  \pi^{1}(a^{1} | s)\pi^{2}(a^{2} | s, a^{1})p(s, (a^{1}, a^{2}), \tilde{s}) \left(\langle \underline{c}(s, a^{1}, a^{2}), \lambda(x)\rangle +  \hat{V}(\tilde{s})\right)\right) - \hat{\beta}, \label{secondDP}
\end{equation}
which has a solution $(\hat{V}, \hat{\beta})$, where $\hat{\beta}$ is unique and $\hat{V}$ is unique upto an additive scalar and can be rendered unique by arbitrarily fixing, say, $\hat{V}(s_{0})=0$. In either (\ref{firstDP}) or (\ref{secondDP}), if we replace $x$ by $x_n$ with $x_n \rightarrow x_{\infty}$, then any subsequential limit $(V'_{\infty}, \beta'_{\infty})$, $(\hat{V}'_{\infty}, \hat{\beta}'_{\infty})$ of the corresponding $(V' = V'_n, \beta' = \beta'_n)$, $(\hat{V}' = \hat{V}'_n, \hat{\beta}' = \hat{\beta}'_n)$ must satisfy the respective dynamic programming equations (\ref{firstDP}),(\ref{secondDP}), with $V'_{\infty}((s_0, a^{1}_{0})) = \hat{V}'_{\infty}(s_{0}) = 0$. By the uniqueness claim above, they are the appropriate value functions for $x = x_{\infty}$. Furthermore, if we pick $\pi^{2}_{n}$ to be a maximizer on the right hand side of (\ref{firstDP}) for $(V'_n, \beta'_n), n \geq 1$, any limit point thereof as $n\uparrow\infty$ must be a maximizer of the same for $n = \infty$. A similar argument works for the minimizers of (\ref{secondDP}). It is easy to deduce from this that the graph of $\chi$ is closed. Hence, $\chi$ is upper semi-continuous. The foregoing also implies that the minmax value is continuous in $x$. This in particular implies that $c^*(\cdot)$ is continuous. The claim regarding $\tilde{\Psi}(\cdot)$ follows from this.

\endproof

\noindent We now recall some  conditions that will ensure the well-posedness of  \eqref{eq:diffincleq1-sg}, along with its proof for sake of completeness.

\begin{proposition}
\label{thm:propn-c(x)-properties-sg}
For each $x \in \mathbb{R}^{K} \setminus D$, the set $w(x)$ is convex,  compact. Also the map $w : x \mapsto w(x)$ is upper semicontinuous.
\end{proposition}

\proof
The set $w(x)$ is clearly bounded from the boundedness assumption on the cost $\underline{c}(\cdot)$.  Also, $w(x)$ is convex by definition. Now, consider the mapping $h(x) :=\{\underline{c}(\tilde{\psi}): \tilde{\psi} \in \tilde{\Psi}(x)\}$. Since $\tilde{\Psi}(x)$ is compact, and $\underline{c}(\cdot)$ is continuous, $h(x)$ is compact. Then, $w(x)$, the closed convex hull of $h(x)$,  is also compact. The upper semicontinuity of $h$ is clear from the upper semicontinuity of $\tilde{\Psi}$. Then the closed convex hull of $h(x)$, i.e., $w(x)$, is also upper semicontinuous, by \cite[Lemma 5, Chapter 5]{Bo08}.

\endproof 


The basic approach to the analysis of \eqref{eq:saeqn1-sg} is to construct a suitable continuous interpolated trajectory $\overline{x}(t), t \geq 0,$ and show that it asymptotically almost surely approaches the solution set of \eqref{eq:diffincleq1-sg}. This is done as follows: Define $t(0) = 0, t(n) = \sum^{n-1}_{m=0} \gamma(m), m \geq 1$. Clearly $t(n) \uparrow \infty$. Let $I_{n} := [t(n), t(n+1)), n \geq 0$. Define a continuous, piecewise linear  $\overline{x}(t), t \geq 0$ by  $\overline{x}(t(n))=x_{n}, n \geq 0,$  with linear interpolation in each interval $I_{n}$. That is,
\begin{equation}
\label{eq:interploated-x(t)}
\overline{x}(t) = x_{n} + (x_{n+1} - x_{n}) \frac{t - t(n)}{t(n+1) - t(n)}, ~ t \in I_{n}.
\end{equation}

Define a $\mathcal{P}(\mathcal{S} \times \mathcal{A})$-valued random process $\mu(t) = [[\mu(t, (s, a))]] , t \geq 0$, by
\begin{equation}
\label{eq:defn-mu-t}
\mu(t, (s,a)) := \delta_{(s_{n}, a_{n})}((s,a)), ~ t \in I_{n}, \ n \geq 0,
\end{equation}
where $\delta$ is  the Kronecker delta function.  Also, define for $t > v \geq 0$, and for  $B$ Borel in $[v, t]$,
\[\mu^{t}_{v}(B \times (s, a)) := \frac{1}{t-v} \int_{B} \mu(y, (s, a)) dy. \]

Two necessary conditions for the analysis in \cite{Bo08} to carry through are:
\begin{enumerate}
\item Almost surely, $\sup_{n} \|x_{n}\| < \infty$.
\item Almost surely, for any $t > 0$, the set $\{\mu^{v+t}_{v}, v \geq 0 \}$ remains tight.
\end{enumerate}

Since we are dealing with only finite spaces, we don't need to worry about the measurability issues discussed in  \cite{Bo08}. For the same reason, conditions 1 and 2 are also true.


For $\nu \in \mathcal{P}(\mathcal{S}\times\mathcal{A})$, define
\[\tilde{h}(x, \nu) := \sum_{(s, a) \in \mathcal{S} \times \mathcal{A}} \left(\underline{c}(s, a) - x \right) \nu(s, a)\]
For $\mu(\cdot)$ defined in \eqref{eq:defn-mu-t}, consider the non-autonomous o.d.e.
\begin{equation}
\label{eq:ode1-mdp}
\dot{x}(t) = \tilde{h}(x(t), \mu(t))
\end{equation}
Let $x^{v}(t), t \geq v$, denote the solution to \eqref{eq:ode1-mdp} with $x^{v}(v) = \overline{x}(v)$, for $v \geq 0$.\\

A set $A \in \mathbb{R}^{K}$ is said to be an \emph{invariant set} for a differential inclusion if for $x(0) \in A$ there is \emph{some} trajectory $x(t), t \in (-\infty, \infty)$ thereof that lies entirely in $A$.  The invariant set is said to be an \emph{internally chain transitive} invariant set  if  for any $x, y \in A$ and any $\epsilon >0, T > 0$, there exists an $n \geq 1$ and points $x_{0}=x, x_{1}, \ldots, x_{n-1}, x_{n}=y$ in $A$, such that some trajectory of the differential inclusion initiated at $x_{i}$ meets with the $\epsilon$-neighborhood of $x_{i+1}$ for $0 \leq i < n$ after a time $t \geq T$. That is, we can reach from any point in the set to any other by traversing finitely many trajectory segments of the differential inclusion interspersed with small jumps. If we saw these through a microscope that cannot detect such jumps, it would appear like a periodic trajectory. This may appear too general, but it does narrow down the possibilities significantly. In fact the whole space is an invariant set for the dynamics, but not necessarily internally chain transitive, indicating in particular that `invariant set' by itself is not a sufficiently informative qualification. Examples of internally chain transitive invariant sets are isolated equilibria, limit cycles, etc. In fact, if $x(\cdot)$ is a bounded trajectory of the differential inclusion, then its $\omega$-limit set $\cap_{t \geq 0}\overline{\{x(s), s \geq t\}}$ (which can depend on the initial condition) is an internally chain transitive invariant set: this follows because the proof of the theorem below also works for the differential inclusion itself. An interesting example of internally chain transitive  invariant sets is the Lorenz attractor  \cite{tucker1999lorenz}. Also, please see Chapter 3 of \cite{barwell2011omega} for a detailed description.  \\

We use the following important result from \cite{BenHofSor1} (see also \cite{Bo08}).
\begin{theorem}\cite[Theorem 4.3]{BenHofSor1}, \cite[Theorem 7, Chapter 6]{Bo08}
\label{thm:thm7-BoCh6}
Under Assumption \ref{assumption1-sg}, almost surely,  \\ $\{\overline{x}(v + \cdot), v \geq 0 \}$ converge to an internally chain transitive invariant set of the differential inclusion
\begin{equation}
\dot{x}(t) \in  \{\tilde{h}(x(t), \nu): \nu \in \Psi(x(t))\} :=  w(x(t)) - x(t) \label{diffinclu}
\end{equation}
as $t\uparrow\infty$. In particular, $\{x_{n}\}$ converge a.s. to such a set.
\end{theorem}

\begin{remark}
The above theorem summarizes the idea of \emph{averaging over natural time-scale} method which we use in the analysis. It shows that  even though the opponent's strategy can be non-stationary, we can restrict our attention to the set of  occupation measures corresponding to the stationary strategies.
\end{remark}

We now show that any path corresponding to the differential inclusion dynamics given by \eqref{eq:diffincleq1-sg}  converges to the set $D$. Some definitions are in order: A compact invariant set $M$ is called an \emph{attractor} of a dynamical system if it has an open neighborhood $O$ such that every trajectory in $O$ remains in $O$ and converges to $M$. The largest such $O$ is called the domain of attraction. An attractor $M$ is called a \emph{global attractor} if the domain of attraction is $\mathbb{R}^{K}$. \\

\noindent {We now give the  proof of Theorem \ref{thm:bat-avggame}:} 

\proof
We construct the interpolated trajectory $\overline{x}(t), t \geq 0$ of \eqref{eq:saeqn1-sg} as defined in equation \eqref{eq:interploated-x(t)}. By Theorem \ref{thm:thm7-BoCh6}, almost surely $\{\overline{x}(v + \cdot), v \geq 0 \}$ converge to an internally chain transitive invariant set of the differential inclusion given by \eqref{eq:diffincleq1-sg}. In particular this implies that $\{x_{n}\}$ converge a.s. to such a set.
Consider the  Lyapunov function $V(x) = \min_{z \in D} \frac{1}{2} \| x - z\|^{2}$.
Since $\nabla V(x) =  (x - P_{D}(x))$, $\frac{d}{dt}V(x(t)) =  \langle \nabla V(x(t)), \dot{x}(t) \rangle =  \langle x(t) - P_{D}(x(t)), {y}(t) \rangle$ for  $y(t) \in w(x(t)) - x(t)$.

 By Proposition \ref{thm:convexityetc} and our hypotheses, there exists an occupation measure $\psi$ such that  $\psi(s,a^{1},a^{2})=\eta^{\psi}(s) \pi^{2}(a^{2}|s,a^{1}) \pi^{1}(a^{1}|s)$, and $c^*(x) =  \langle \underline{c}(\psi), \lambda(x)\rangle  \leq \langle P_{D}(x), \lambda(x)\rangle$.  Then for any  policy $\tilde{\psi}_{x} \in \tilde{\Psi}(x)$ (which can be chosen to be measurable in $x$ by a standard measurable selection theorem \cite{Wagner}), we have $\langle \underline{c}(\tilde{\psi}_{x}), \lambda(x)\rangle \leq c^{*}(x) \leq \langle P_{D}(x), \lambda(x)\rangle$.  So for any $x(\cdot)$ satisfying (\ref{diffinclu}),
 \[\langle \underline{c}(\tilde{\psi}_{x(t)})-P_{D}(x(t)), x(t) - P_{D}(x(t) \rangle  \leq 0~~\text{for all}~~\tilde{\psi}_{x(\cdot)} \in  \tilde{\Psi}(x(\cdot))\]
and hence $\langle \underline{c}(\tilde{\psi}_{x(t)})- x(t), x(t) - P_{D}(x(t) \rangle \leq - \|x(t) - P_{D}(x(t))\|^{2}$. This gives
\[\frac{d}{dt}V(x(t)) \leq -2 V(x(t)) \]
so that
\[V(x(t)) \leq V(x(0)) e^{-2t}.\]
Thus $D$ is a global attractor. Then the  internally chain invariant set  corresponding to the differential inclusion \eqref{eq:diffincleq1-sg} is a  subset of D \cite{Bo08} . Hence, $\{x_{n}\}$ converges to $D$ a.s., proving the first claim. The proof of the second claim is also implied by the foregoing.

\endproof

We complete the characterization by specifying the necessary conditions for approachability. 
\begin{proposition}[Necessary Condition]
\label{prop:necess-sg}
If a closed convex set $D$ is approachable from all initial states in an arbitrary Stackelberg stochastic game  satisfying Assumption \ref{assumption1-sg}, then\\
(i) every half-space containing $D$ is approachable, and\\
(ii) $\forall x$, there exists an occupation measure $\psi$, $\psi(s,a^{1},a^{2})=\eta^{\psi}(s) \pi^{2}(a^{2}|s, a^{1}) \pi^{1}(a^{1}|s)$,  such that  $\underline{c}(\psi) = c^*(x)$ and   $\langle \underline{c}(\psi) - P_{D}(x) , \lambda(x)\rangle \leq 0$.
\end{proposition}
\proof
Claim (i) is obvious. We now show that (i) implies (ii) and complete the argument.

Let $x \in \mathbb{R}^{K} \setminus D$ and $H_{x}$ be the supporting half-space to the set $D$ at the point $P_{D}(x)$ given by
\[H_{x} := \{y \in \mathbb{R}^{K}: \langle y - P_{D}(x), \lambda(x) \rangle \leq 0\}.\]
Since every half-space containing $D$ is approachable, there exists a  strategy $\sigma^{1}$ for agent $1$ such that for any (Stackelberg) strategy $\sigma^{2}$ of agent $2$, $\lim \sup_{n \rightarrow \infty} \langle x_{n} - P_{D}(x), \lambda(x) \rangle \leq 0$, $\mu(\sigma^{1}, \sigma^{2})$ a.s. Since  $|\langle x_{n} - P_{D}(x), \lambda(x) \rangle|$  is bounded,
\begin{equation}
\label{eq:nec-sg-st1}
 \inf_{\sigma^1} \sup_{\sigma^2} \lim \sup_{n \rightarrow \infty} \mathbb{E}_{\mu(\sigma^1, \sigma^2)}[ \langle x_n, \lambda(x) \rangle ]\leq \inf_{\sigma^{1}}  \sup_{\sigma^{2} }\mathbb{E}_{\mu(\sigma^{1}, \sigma^{2})}[\lim \sup_{n \rightarrow \infty} \langle x_{n}, \lambda(x) \rangle]  \leq \langle P_{D}(x), \lambda(x) \rangle
\end{equation}
Note that the L.H.S. of equation \eqref{eq:nec-sg-st1} is the min-max cost for agent $1$ in the  average cost scalar-valued Stackelberg stochastic game with stage cost $\tilde{c}(s, a;x)=\langle  \underline{c}(s, a), \lambda(x) \rangle$. By the arguments in the proof of Proposition \ref{thm:convexityetc}, the min-max cost is achieved by a pair of stationary Markov policies $(\pi^1(\cdot | \cdot), \pi^2(\cdot | \cdot, \cdot))$ obtained by solving a pair of dynamic programs, implying in particular that there exists a $\psi \in \Psi(x)$ such that the L.H.S. is equal to   $\langle \underline{c}(\psi), \lambda(x) \rangle$. This completes the proof.

\endproof

\begin{remark}
The differential inclusion trajectory approaches $D$ at an exponential rate. There is, however, a time scaling $n \mapsto t(n)$ in our passage from the original iterates to the differential inclusion approximation. Since $t(n) = \sum_{m=0}^n\gamma_1(m)$ and $\gamma_1(n) = \frac{1}{n+1}$, $t(n) \approx \log n$. Thus exponential decay translates into an inverse power law decay in the original time scale. There is a further issue of discretization errors and errors due to noise. As shown in \cite{Bo08}, Chapter 4, the interpolated iterates $\bar{x}(\cdot)$ remain within a small tube a differential inclusion trajectory after $n \geq n_0$ steps\footnote{The estimates of \textit{ibid.} are for an o.d.e.\ limit, but a similar argument works for differential inclusion limits.} with probability exceeding
\begin{equation}
1 - O(\left(e^{-\frac{C}{\sum_{m=n_0}^{\infty}\gamma_1(m)^2}}\right) \label{err}
\end{equation}
for some constant $C > 0$, which for $\gamma_1(n) := \frac{1}{n+1}$ becomes an exponential decay in $n$.
\end{remark}

\section{Extensions}

\subsection{Extension to Non-Convex Sets}
\label{sec:non-convex-sg}

We now give the approachability result when the target set is non-convex. The proof is the same as that of Theorem \ref{thm:bat-avggame} except for the fact that the Lyapunov function may be non-differentiable. We overcome this difficulty by considering semidifferentials and a general version of  the \emph{envelope theorem} \cite{bardi2008optimal}. We assume that Assumption \ref{ass:unichain} is true throughout this section.

We first give some  some basic definitions and results from \cite[Pages 29, 42-46]{bardi2008optimal} that are required for the proof.

\begin{definition}[Semi-differentials]
Let $V : \mathbb{R}^{K} \rightarrow \mathbb{R}$. The \emph{super} and \emph{sub}-differential of $V$ (or \emph{semi-differentials}) at $x$, $D^{+}V(x)$ and $D^{-}V(x)$, are defined as
\begin{align*}
D^{+}V(x) &:= \left\{p \in \mathbb{R}^{K}: \limsup_{y \rightarrow x}  \frac{V(y)-V(x)-p \cdot (y-x)}{|x-y|} \leq 0 \right\} \\
D^{-}V(x) &:= \left\{p \in \mathbb{R}^{K}: \liminf_{y \rightarrow x}  \frac{V(y)-V(x)-p \cdot (y-x)}{|x-y|} \geq 0 \right\}
\end{align*}
\end{definition}

Let  $V$ be such that
\[V(x) := \inf_{z \in D} g(x, z) \]
where $g : \mathbb{R}^{K} \times D \rightarrow \mathbb{R}$, $D \subset \mathbb{R}^{K}$. Assume that,
\begin{align}
\label{eq:gb-assumptions1}
 &\text{(A1)}~~ g~\text{is bounded and}~ g(\cdot, z)~\text{differentiable at $x$ uniformly in $z$}, \\
\label{eq:gb-assumptions3}
&\text{(A2)}~~  z \longmapsto D_{x}g(x, z)~\text{is continuous and}~ z \longmapsto g(x, z)~\text{lower semicontinous,}
\end{align}
where $D_{x}g$ is the partial derivative of $g$ w.r.t. $x$. Let
\[\tilde{P}_{D}(x) := \arg \min_{z \in D} g(x, z) := \{z \in D : V(x) = g(x, z) \},\hspace{1cm} Y(x) := \{D_{x}g(x, z) : z \in \tilde{P}_{D}(x)\}\]
Also, the (one-sided) \emph{directional derivative}  of $V$ in the direction of $q$ is
\[ D^{+}V(x)(q): = \lim_{h \rightarrow 0^{+}} \frac{V(x+hq) - V(x)}{h}. \]
We use the following general version of the \emph{envelope theorem}.
\begin{proposition}
\label{subsuper}\cite[Proposition 2.13, Page 44]{bardi2008optimal}
Let $D$ be a compact set and $g$ satisfies assumptions \eqref{eq:gb-assumptions1}-\eqref{eq:gb-assumptions3}. Then,
\begin{align*}
Y(x) &\neq \emptyset \\
D^{+}V(x) &= \overline{co}Y(x)\\
D^{-}V(x) &= \left \{ \begin{array}{cc}
\{y\} \quad & \text{if}~ Y(x)=\{y\} \\
\emptyset \quad &~\text{if $Y(x)$ is not a singleton}
\end{array} \right.
\end{align*}
Moreover, $V$ has the (one-sided) directional derivative in any direction $q$, given by
\[D^{+}V(x)(q) = \min_{y \in Y(x)} y \cdot q = \min_{p \in D^{+}V(x)} p \cdot q \]
\end{proposition}

Let $g(x, z) = \|x - z\|^2$. Then $\tilde{P}_{D}(x)$ is the set of points in $D$ that are closest to $x \in \mathbb{R}^{K} \setminus D$.
\begin{theorem}
\label{thm:bat-avggame-nonconvex}
(i) (Sufficient Condition) A closed  set $D$ is approachable from all initial states in the  stochastic game satisfying Assumption \ref{assumption1-sg} if for every $x \in \mathbb{R}^{K} \setminus D$ and  for each $P_{D}(x) \in \tilde{P}_{D}(x)$, there exists an  occupation measure $\psi$, $\psi(s,a^{1},a^{2})=\eta^{\psi}(s) \pi^{2}(a^{2}|s,a^{1}) \pi^{1}(a^{1}|s)$, such that  $\underline{c}(\psi) = c^*(x)$ and   $\langle \underline{c}(\psi) - P_{D}(x) , \lambda(x)\rangle \leq 0$.  \\
(ii) (Strategy for Approachability)  A strategy for approachability is:  At every time instant $n$,  select a $P_{D}(x) \in \tilde{P}_{D}(x)$, and select action $a^{1}_{n}$ according to the policy $\pi^{1}(\cdot | x_{n})$ such that $\pi^{1}(\cdot | x_{n}) \in \Pi^{1}(x_{n})$.
\end{theorem}

\proof
Let  $V(x) = \min_{z \in D} \frac{1}{2} \| x - z\|^{2}$ be the Lyapunov function. Then by Proposition \ref{subsuper},
\begin{displaymath}
D^{+}V(x)=\overline{co}\{(x-P_{D}(x)) : P_{D}(x) \in \tilde{P}_{D}(x) \}.
\end{displaymath}
 Let $v(t) = V(x(t))$ and $D^+v(t)$ its right derivative. By chain rule \cite[Proposition 7, Page 288]{aubin1984differential},
\[D^{+}v(t)=D^{+}V(x(t))(\dot{x}(t)) = D^{+}V(x(t))(y(t))~~~\text{for some}~  y(t) \in w(x(t)) - x(t).\]
As before, we can also show that for all  $P_{D}(x(t)) \in \tilde{P}_{D}(x(t))$,
\[\langle \underline{c}(\tilde{\psi}_{x(t)})- x(t), x(t) - P_{D}(x(t)) \rangle \leq - \|x(t) - P_{D}(x(t))\|^{2} = -2 V(x(t)) = -2 v(t).\]
Then,
\[D^{+}v(t) = \min_{P_{D}(x(t)) \in \tilde{P}_{D}(x(t))} \langle \underline{c}(\tilde{\psi}_{x(t)}) - x(t), x(t) - P_{D}(x(t) \rangle \leq -2 v(t) \]
and by \cite[Proposition 8, Page 289]{aubin1984differential}, we get
\[v(t) - v(0) + 2 \int^{t}_{0} v(\tau) d\tau \leq 0. \]
By Gronwall inequality,
\[v(t) \leq v(0) e^{-2t}.   \]
The rest of the proof is the same as in the proof of Theorem \ref{thm:bat-avggame}.

 \endproof 

 \subsection{Perfect Information Stochastic Games}
We can extend our results to a special class of simultaneous move games called perfect information stochastic games defined as follows.
\begin{definition}
Let $\mathcal{A}^{i}(s)$ be the set of feasible actions that the agent $i$ can select when the state is $s$. A stochastic game is said to be a game of perfect information if the state space $\mathcal{S}$ can be partitioned into two disjoint sets $\mathcal{S}^{1}$ and $\mathcal{S}^{2}$ such that  $|\mathcal{A}^{2}(s)| = 1$ for $s \in \mathcal{S}^{1}$ and $|\mathcal{A}^{1}(s)| = 1$ for $s \in \mathcal{S}^{2}$.
\end{definition}
\noindent In words, at any given state $s$ at least one of the agents has only a single action available.

It is easy to see that the quantities defined in equations  \eqref{eq:minmaxcost-sg}-\eqref{eq:tildaPsi(x)-sg} for a Stackelberg stochastic game are well defined for a perfect information simultaneous move stochastic game. Note that the only change will be in the way we define the adversary's (agent 2) strategy. In the Stackelberg game, it is denoted as $\pi^{2}(a^{2}|s, a^{1})$. However, in a perfect information game, $\pi^{2}(a^{2}|s, a^{1}) = \pi^{2}(a^{2}|s)$ for $s \in \mathcal{S}^{2}$ and $\pi^{2}(a^{2}|s, a^{1}) = \delta_{a_{0}}(a^{2})$ for $s \in \mathcal{S}^{1}$ where $a_{0}$ is the single action available to agent $2$ when $s \in \mathcal{S}^{1}$.

In  Section \ref{sec:cond-suff-necc} we explicitly used the Stackleberg game assumption mainly to prove Proposition  \ref{thm:convexityetc}. We needed that assumption to decouple the dynamics as a combination of a zero-sum game and an MDP. In particular, the proof of the proposition starts by observing that the adversary (agent $2$) faces an average reward MDP. It is easy to see that this observation is true in the case of a perfect information simultaneous move stochastic game as well.  When the state $s \in \mathcal{S}^{1}$, the adversary's decision is trivial: select action $a_{0}$. When the state $s \in \mathcal{S}^{2}$, it faces a completely observed MDP because the action of agent $1$ is known (since the action space is a singleton). So, the proof of Proposition \ref{thm:convexityetc} will carry through.  The proof of Theorem \ref{thm:bat-avggame} depends on the Stackleberg assumption only via Proposition \ref{thm:convexityetc}.  So, our main result (Theorem \ref{thm:bat-avggame}) extends naturally to the case of a perfect information simultaneous move stochastic game.

However, it is not immediately clear if our approach can be extended to a switching-controller stochastic game. Switching-controller stochastic game is a generalized version of perfect information stochastic games. The transition kernels are such that when the state $s \in \mathcal{S}^{i}$, only agent $i$ can influence the transition. But the other agent's action set need not be a singleton. So the adversary is facing a partially observed MDP when $s \in \mathcal{S}^{2}$. Also, its decision making when the state $s \in \mathcal{S}^{1}$  is not trivial because the action set is not a singleton. So, the proof of Proposition  \ref{thm:convexityetc} doesn't carry through directly. We plan to address this issue in a future work.

 \section{A Learning Algorithm for Approachability in Stochastic Stackelberg Games}
 \label{sec:learning}
Approachability theorem for Stackelberg stochastic  games that we proved above shows that if  agent $1$ selects her action at time step $n$ according to the policy $\pi^{1}(\cdot | x_{n})$ such that $\pi^{1}(\cdot |x_{n}) \in \Pi^{1}(x_{n})$, then $x_{n}$ approaches the desired set $D$. Given $x_{n}$, such a policy can be easily computed if one knows the transition kernel $p(\cdot, \cdot, \cdot)$. The problem of `learning' arises when this transition kernel is unknown and the  objective of the learning  algorithm is to `learn' such a policy $\pi^1(\cdot |x_{n})$ at each time step $n$. We give an algorithm  which  indeed does this. We do this by considering the problem as learning in two coupled MDPs.  We assume that Assumption \ref{assumption1-sg}  is true throughout this section. 

The proposed algorithm below is a two time scale stochastic approximation algorithm. We use two step-size sequences $\gamma_{1}(n)$ and $\gamma_{2}(n)$.  
The step-sizes $\gamma(n) = \gamma_{1}(n), \gamma_{2}(n)$ satisfy the conditions
\begin{align}
\label{eq:conditions-gamma}
\sum_{n} \gamma(n) = \infty, &\quad \sum_{n} \gamma^{2}(n) < \infty \\
\label{eq:condition-asynchronous}
\sup_{n} \frac{\gamma([yn])}{\gamma(n)} < \infty, &\quad \frac{\sum^{[yn]}_{m=0}\gamma(m)}{\sum^{n}_{m=0}\gamma(m)} \rightarrow 1 ~\text{uniformly in}~y \in (0, 1)
\end{align}
with the additional stipulation that $\gamma_1(n) = o(\gamma_2(n)).$

We give the asynchronous learning algorithm below. The synchronous scheme may be written analogously.  Note that in the following $\{a^{1}_{n}\}$ denotes the actual control actions of agent 1 while $\{a^{2}_{n}\}$ denotes the \textit{simulated} control actions of agent 2. The actual control actions of agent 2 is denoted by $\{\bar{a}^{2}_{n}\}$. The reason for this demarcation is explained later.

\noindent \textbf{Learning Algorithm:} 
\begin{enumerate}
\item Initialize $n=0$. Initialize the $Q$-values $\widehat{Q}_{0}(s, a^{1}, a^{2}) =   \widetilde{Q}_{0}(s, a^{1}, a^{2}) = 0, \forall s, a^{1}, a^{2}$. \\ Set average cost vector $x_{0} = 0$.  Observe the state $s_{0}$.  \\ Select an action $a^{1}_{0} \in \mathcal{A}^{1}$ uniformly at random.  \\
Agent 2 takes action $\bar{a}^{2}_{0}$.  Observe the cost $\underline{c}(s_{0}, a^{1}_{0}, \bar{a}^{2}_{0})$ 
\item Update the time step $n \leftarrow n+1$
\item Observe $s_{n+1}$. Simulate $\hat{a}^{1}_{n+1}$ according to the distribution $\pi^{\epsilon, 1}_{n}$ (c.f. equation \eqref{eps}). \\ Update the $Q$-value  $\widehat{Q}_{n+1}(s, a^{1}, a^{2})$ according to \eqref{eq:qn-learning2-sg}
\item Update the $Q$-value  $\widetilde{Q}_{n+1}(s, a^{1}, a^{2})$ according to \eqref{eq:qn-learning1-sg}  
\item Choose  action $a^{1}_{n+1}$ according to the policy $\pi^{\epsilon, 1}_{n+1}$ as defined in  \eqref{eps}
\item Simulate action $a^{2}_{n+1}$ according to the policy $\pi^{\epsilon, 2}_{n+1}$ as defined in  \eqref{eps}
\item Agent 2 takes  action $\bar{a}^{2}_{n+1}$. Observe the cost $\underline{c}(s_{n+1},  a^{1}_{n+1}, \bar{a}^{2}_{n+1} )$
\item Update the average cost according to  \eqref{eq:xn-learning-sg}
\item Update $\hat{\nu}(n+1, s, a^{1}, a^2)$ according to  \eqref{eq:nu-defn}
\item Return to step 2
\end{enumerate}

\begin{align}
\label{eq:xn-learning-sg}
x_{n+1} &= x_{n} + \gamma_{1}(n) \left(\underline{c}(s_{n+1},  a^{1}_{n+1}, \bar{a}^{2}_{n+1}) - x_{n}\right), \gamma_{1}(n) = 1/(n+1),\\
\label{epdecay}
\epsilon_{n+1} &= \epsilon_n(1 - \gamma_1(n)),\\
\label{eq:qn-learning2-sg}
\widehat{Q}_{n+1}((s, a^{1}), a^{2}) &= \widehat{Q}_{n}((s, a^{1}), a^{2}) + \gamma_{2}(\hat{\nu}(n,s,a^{1},a^2)) I\{((s_{n}, a^1_n), a^2_{n}) = ((s, a^{1}), a^{2})\} \nonumber\\
&\big ( \tilde{c}((s, a^{1}), a^{2}; x_{n}) +  \max_z \widehat{Q}_{n}((s_{n+1}, \hat{a}^1_{n+1}), z)  - f(\widehat{Q}_{n}) - \widehat{Q}_{n}((s, a^{1}), a^2) \big),  \\
\label{eq:qn-learning1-sg}
\widetilde{Q}_{n+1}(s, (a^{1}, a^{2})) &= \widetilde{Q}_{n}(s, (a^{1},a^{2})) + \gamma_{2}(\hat{\nu}(n, s, a^{1}, a^{2})) I\{s_{n} = s, a^1_n = a^{1}\} \nonumber\\
& \times \big ( \tilde{c}(s, a^{1}_n, a^{2}_n; x_{n}) +  \min_y \widetilde{Q}_{n}(s_{n+1}, (y, \arg \max_{z}  \widehat{Q}_{n+1}(s, y, z))) \nonumber \\
 & - \ f(\widetilde{Q}_{n}) - \widetilde{Q}_{n}(s, (a^{1}, a^{2})) \big),  \\
\label{eq:policy-learning1-sg}
\pi_{n+1}^1(a^{1} | s) &= \delta_{a^*(1)}(a^{1}),  \ \mbox{where}  \\
a^*(1) &:= \arg \min_{a^{1}} \widetilde{Q}_{n+1}(s, (a^{1}, \arg \max_{a^{2}}  \widehat{Q}_{n+1}(s, a^{1}, a^{2}))). \nonumber \\
\pi_{n+1}^2(a^{2} | (s, a^{1})) &= \delta_{(a^*(1), a^{*}(2))}((a^{1}, a^{2})),  \ \mbox{where}  \\
a^*(2) &:= \arg \max_{a^{2}}  \widehat{Q}_{n+1}(s,  a^{*}(1), a^2). \nonumber 
\end{align}
Here the actual control process $\{a^1_n\}$ for agent 1 and \textit{simulated} control actions $\{a^2_n\}$ for agent 2  are governed by the randomized Markov policy 
\begin{equation}
\pi^{\epsilon, i}_{n+1}(\cdot) = (1 - \epsilon_n)\pi_{n+1}^{i}(\cdot) + \epsilon_n \mu^{i}(\cdot),  \label{eps}
\end{equation}
where $\mu^{i} :=$ the uniform distribution on $\mathcal{A}^{i}, i = 1, 2,$.

The function  $f$ is any Lipschitz function satisfying the following properties: For the all $1$ vector $e$, $f(e)=1$, $f(y+ce)=f(y)+c$  and $f(cy) = c f(y)$ for $c \in \mathcal{R}$. As a simple example, we can set $f(Q) = Q(s_{0}, a_{0})$ for a fixed $s_{0} \in \mathcal{S}, a_{0} \in \mathcal{A}$. Other possible choices are maximum, minimum or arithmetic mean of $\{Q(i,s)\}$. Also, 
\begin{equation}
\label{eq:nu-defn}
\hat{\nu}(n, s, a^{1}, a^{2}) = \sum^{n}_{m=1} I\{(s, a^{1}, a^{2}) = (s_{m}, a^{1}_{m}, a^{2}_{m})\}.
\end{equation}

The synchronous scheme is analogous except that all components of $\{\widehat{Q}_n\}, \{\widetilde{Q}_n\}$ are updated simultaneously. Thus the indicators $I\{((s_n,a^1_n),a^2_n) = ((s,a^1),a^2)\}$ in (\ref{eq:qn-learning2-sg}) and $I\{s_n = s, a^1_n = a\}$ in (\ref{eq:qn-learning1-sg}) drop out and the stepsizes $\gamma_2(\hat{\nu}(n,s,a^1,a^2))$ are replaced by $\gamma_2(n)$.

\begin{assumption}
Relative sampling frequency of state-action pairs  is bounded away from zero, i.e.,
\begin{equation}
\label{eq:relsamplingfreq}
\liminf_{n \rightarrow \infty} \frac{\hat{\nu}(n,s, a^{1}, a^{2})}{n+1} > 0 \ ~\text{a.s.}~ \ \forall \ (s, (a^{1}, a^{2})) \in \mathcal{S} \times \mathcal{A}.
\end{equation}
\end{assumption}

This is not ensured automatically by the above choice of $\pi^{1}_{n+1}$, so one usually employs some randomization to ensure adequate `exploration' implicit in the condition \eqref{eq:relsamplingfreq}. One way to do so is to choose some $0 < \epsilon << 1$ and use instead $\pi^{\epsilon, 1}_{n+1}(s, \cdot)$ defined in (\ref{eps}) with $\epsilon_n \equiv \epsilon > 0$ and
$\mu$ is uniform on $\mathcal{A}^{1}$. This will ensure that any combination $(s, a^1)$ will have relative frequency bounded away from zero. Our slow decrease of $\epsilon_n$ will effectively achieve the same effect by the `two time scale' argument as we see later.

Our requirement, however, is for the corresponding statement for all triplets $(s, a^1, a^2)$, not just for $a^1$. The control sequence $\{a^2_n\}$ is chosen by agent 2, or `nature', so we cannot ordain this, nor can we suppose that agent 2 run the Q-learning algorithm stipulated above. We can, however, ensure this by changing the interpretation of the algorithm, which we describe next.  We use the terms `adversary', `nature' and `agent 2' interchangeably.

That `nature' choose controls to satisfy (\ref{eq:relsamplingfreq}) is not required at all if we view (\ref{eq:qn-learning2-sg}) and subsequent choice of $a^*(2)$ that follows as agent 1's \textit{simulation} of the worst case behavior of agent 2. Thus this becomes an `off-line' algorithm for agent 2. We briefly recall what this distinction means.
In Q-learning, the minimizer of the estimated Q function over the control variable gives an estimate of the optimal control choice at a given state (which forms the first argument of the Q function). There is no requirement, however, that the actual simulation be carried out according to the policy suggested by this choice. It can be any admissible state-control process as long as (\ref{eq:relsamplingfreq}) is satisfied. This is the \textit{off-line} situation in contrast with the \textit{online} case wherein $\{(s_n, a_n)\}$ is not a simulation, but an actual run of the control system, so the control choice matters. Here we are perforce considering an on-line situation for agent 1 who wants to drive the averaged vector cost to a target set, but s/he is free to simulate the adversary's behavior assuming the worst case, so that the latter becomes an off-line simulation. Therefore while (\ref{eq:relsamplingfreq})  imposes restrictions on agent 1's control choices as noted above,  there can be no a priori restriction on agent 2's control choices because they are only simulated and therefore can be chosen at will, in particular so as to satisfy (\ref{eq:relsamplingfreq}). Note, however, that the iterates (\ref{eq:xn-learning-sg}) must be performed according to actual observed costs.

 That is, the adversary's control choices entering (\ref{eq:xn-learning-sg}) are the actual control choices of the adversary so that $\underline{c}(s_{n+1},\bar{a}_{n+1})$ is the true cost observed, whereas the control choices of the adversary entering (\ref{eq:qn-learning2-sg})-(\ref{eq:qn-learning1-sg}) are the simulated ones as stated.   We have  demarcated  this by using the notation $\bar{a}_{n} = (a^{1}_{n},  \bar{a}^{2}_{n})$ for actual controls used as opposed to $a_n = (a^{1}_{n}, a^{2}_{n})$ for the simulated ones. Of course, the first component of these is common.

\begin{remark}
It is also possible that we do not explicitly know the function $\underline{c}(\cdot, \cdot)$, but observe a noisy version thereof, say,
$\breve{c}(n) = \underbar{c}(x_n, \bar{a}_{n}) \ +$ a zero mean independent noise variable with finite second moment. Then we replace $\underline{c}(s_{n+1}, \bar{a}_{n+1})$ in (\ref{eq:xn-learning-sg}) by this noisy observation. We need not do so for the occurrences of $\underline{c}, \tilde{c}$ in the rest of the algorithm, because there we actually need to plug in the function value at the appropriate state-control combination, which is known because the function $\underline{c}$ is known. The zero mean noise in observed cost can be absorbed into the `martingale difference noise' term $M_{n+1}$ in (\ref{mg}) for (\ref{eq:xn-learning-sg}). This does not affect the stochastic approximation based analysis thereof.

\end{remark}

\begin{remark} The recent work of Yu and Bertsekas \cite{YuBertsekasSSP} gives an elegant and highly intricate argument in case of single agent Q-learning for stochastic shortest path problems that relaxes (\ref{eq:relsamplingfreq}) to the weaker requirement that $\tilde{\nu}(n,s,a^{1}) \uparrow \infty$ a.s. A similar result should be possible in the present context, which will make the requirement (\ref{eq:relsamplingfreq}) unnecessary.
\end{remark}

\noindent \textbf{Explanation of the learning algorithm:} 

The explanation for these iterations is as follows, which also gives an intuition about the convergence proof to follow. As in classical Blackwell approachability, agent 1 hedges against the worst possible behavior of the adversary, i.e., agent 2, whose exact strategy agent 1 may not know. This reduces it to a position-dependent zero-sum game in the following manner. Given the present value of the vector average cost, agent 1 wants to maximize its drift towards the target set. If she hedges against the possibility of agent 2 minimizing the same, she ensures movement towards the target set in the worst possible scenario.  Thus we have the MDPs associated with agent 1 and 2 as in the proof of Proposition 1. 

If we fix $x_n \equiv x$ and $\pi^1_n \equiv \pi^1$, agent 2 faces the MDP: Control the chain $\{(s_n, a^1_n)\}$ with controlled transition probabilities
$$\tilde{p}((\tilde{s}, \tilde{a}^1) | (s, a^1), a^2) := p(s, (a^1, a^2), \tilde{s})\pi^1(\tilde{a}^1|\tilde{s})$$
with control variable $a^2$, so as to maximize the ergodic cost
\begin{equation}
\lim_{n\uparrow\infty}\frac{1}{n}\sum_{m = 0}^{n - 1}E[\langle c(s_m, a^1_m, a^2_m), \lambda(x)\rangle]. \label{qlearnfirst}
\end{equation}
(We use `$\lim$' in place of the usual `$\liminf$' because of our prior restriction to stationary strategies justified in section 2.) The dynamic programming equation for this control problem was noted to be (\ref{firstDP}). Denoting the term in large parentheses on the right hand side of (\ref{firstDP}) as $\check{Q}((s,a^1), a^2)$, we have the corresponding equation for `Q-values' $\check{Q}((\cdot, \cdot), \cdot)$ as
\begin{equation}
\check{Q}((s,a^{1}), a^2) = \sum_{(\tilde{s}, \tilde{a}^{1})}\tilde{p}((\tilde{s}, \tilde{a}^{1}) | (s, a^{1}), a^{2}) \left(\langle c(s, a^{1}, a^{2}), \lambda(x)\rangle + \max_{b}\check{Q}((\tilde{s}, \tilde{a}^{1}), b) \right) - \beta. \label{firstQP}
\end{equation}
This has the advantage compared to (\ref{firstDP}) of having the nonlinearity, i.e., the max operator,  \textit{inside} the conditional expectation, i.e., averaging w.r.t.\ $\tilde{p}$. This allows us to write a stochastic approximation scheme to solve the fixed point equation (\ref{firstQP}) by first replacing the conditional average by an evaluation at a random variable realized according to $\tilde{p}$ and then making an incremental correction in the left hand side based on the difference between the right hand side and the left hand side. The incremental corrections with suitably chosen weights lead to a stochastic approximation scheme which does the conditional averaging for you. This will be
\begin{eqnarray}
\check{Q}_{n+1}((s, a^{1}), a^{2}) &=& \check{Q}_{n}((s, a^{1}), a^{2}) + \gamma_{2}(\hat{\nu}(n,s,a^{1},a^2)) I\{((s_{n}, a^1_n), a^2_{n}) = ((s, a^{1}), a^{2})\} \nonumber \\
&&\times \big ( \tilde{c}((s, a^{1}), a^{2}; x) +  \max_z \check{Q}_{n}((s_{n+1}, a^1_{n+1}), z)  - f(\check{Q}_{n}) - \check{Q}_{n}((s, a^{1}), a^2) \big). \nonumber\\
 \ && \ \label{checkQ}
\end{eqnarray}
Here $f(\check{Q}_n)$ is a surrogate for $\beta$ which is also unknown a priori and gets specified by (\ref{firstQP}) as the optimal reward. If we generate samples according to $\tilde{p}$ by a single simulation run of a Markov chain $\{(s_n, a^1_n)\}$ governed by $\tilde{p}$, we see only the transition out of the current state at a given time and therefore can update only the component corresponding to the current state. This is incorporated in the above by means of the indicator $I\{((s_{n}, a^1_n), a^2_{n}) = ((s, a^{1}), a^{2})\}$, which makes this an \textit{asynchronous} stochastic approximation. This is the celebrated Q-learning algorithm. Our iteration \eqref{eq:qn-learning2-sg} is exactly the same as (\ref{checkQ}) above  with $x_n \equiv x, \ \pi^1_n \equiv \pi^1$.
Thus in the present framework,  $\widehat{Q}_{n}$ replace $\check{Q}_n$ and  are the estimates for  $Q$-function for the MDP faced by agent $2$.
 $\widetilde{Q}_{n}$ are estimates for the $Q$-function for the MDP faced by agent $1$ and ${\pi}^{1}_{n}$ are estimates for its strategy.
Both $x_{n}$ and $\epsilon_n$ are updated on a slower `algorithmic' timescale compared to $\widehat{Q}_{n}$ because of our conditions $\gamma_2(n) = o(\gamma_1(n))$, and therefore can be considered `quasi-static' for the analysis of \eqref{eq:qn-learning2-sg} by the theory of `two time scale stochastic approximation' as we shall see below.  That is, we can treat them as constant for analyzing \eqref{eq:qn-learning2-sg}. This will allow us to conclude that $\widehat{Q}_{n}$  asymptotically track the optimal Q-value for agent 2's MDP parametrized by the slowly varying $x_{n}, \epsilon_n$. To be precise,
\begin{equation}
\label{eq:tracking}
\|\delta_{\arg \max_{z}  \widehat{Q}_{n}(\cdot, \cdot, z)} - \Gamma_{\pi^1_n}(x_n)\| \rightarrow 0 \ \ \mbox{a.s.}
\end{equation}
Recall that the dependence on $\epsilon_n$ is via the actual control policy implemented, see (\ref{qlearnfirst}) and the discussion below.

Consider \eqref{eq:qn-learning1-sg} next. Again, $x_n, \epsilon_n$ can be treated as quasi-static for this and \eqref{eq:tracking} holds. Then the problem faced by agent 1 is to minimize
\begin{displaymath}
\max_{\pi^2}\lim_{n\uparrow\infty}\frac{1}{n}\sum_{m = 0}^{n - 1}E[\langle c(s_m, a^1_m, a^2_m), \lambda(x)\rangle].
\end{displaymath}
Thus agent 1  faces the dynamic programming equation  \eqref{secondDP} where $\pi^2$ is chosen to be optimal for agent 2. Equation \eqref{eq:qn-learning1-sg} then represents the Q-learning algorithm for agent $1$ with $\delta_{\arg \max_z\widehat{Q}_n(\cdot, \cdot, z)}$ serving as the estimate for optimal $\pi^2$, justified by \eqref{eq:tracking}.

Our main convergence theorem is the following: 

\begin{theorem}
\label{thm:learning-sg}
For the learning algorithm for approachability in Stackelberg stochastic game given by equations \eqref{eq:xn-learning-sg} - \eqref{eq:policy-learning1-sg}, $\|x_{n} - D\| \rightarrow 0$ almost surely.
\end{theorem}

\proof 

The proof progresses in several steps. Since it uses known ideas from previous work in the context of Markov decision processes, we shall be brief with details where they are lengthy but essentially the same as the MDP case, giving precise pointers to the literature.
\begin{itemize}

\item \textit{Step 1: Convergence of $\{\widehat{Q}_n\}$:}

We first argue that $\{\widehat{Q}_n\}$  converge a.s.\ assuming that they remain bounded a.s., which we prove subsequently. The proof is based on the `o.d.e.\ approach' for analysis of stochastic approximation algorithms. We first consider the corresponding synchronous scheme. Since $\gamma_1(n) = o(\gamma_2(n))$, (\ref{eq:xn-learning-sg}) is on a slower   time scale than (\ref{eq:qn-learning2-sg})  and we analyze (\ref{eq:qn-learning2-sg}) treating $x_n \approx x$ as frozen by the two time scale argument of section 6.1, \cite{Bo08}. Likewise we treat $\epsilon_n \approx \epsilon$ as frozen as it varies on the same time scale as $\{x_n\}$.  The limiting o.d.e.\ for (\ref{eq:qn-learning2-sg}) then is
    \begin{equation}
    \dot{q}(t) = F^{x}(q(t)) - q(t), \label{odelim1}
    \end{equation}
    where components of $q(\cdot), F^{x}(\cdot)$ are indexed by $(s,a^1,a^2)$ listed lexicographically with
\begin{eqnarray}
 \label{asynlambda}
  F^{x}_{(s,a^1,a^2)}(q) &:=&  \tilde{c}((s, a^1), a^2; x) - f(q) - q((s,a^1),a^2)  + \nonumber \\ &&    \sum_{(\tilde{s},\tilde{a}^1)}\tilde{p}((\tilde{s},\tilde{a}^1)|(s,a^1),a^2)\max_{a'}q((s,a^1),a').
\end{eqnarray}
    The convergence of this to the unique fixed point of $F^x$ follows exactly as in Theorem 3.5 of \cite{Abounadi2001learning}. The boundedness proof also follows essentially as in \textit{ibid.} using the arguments of \cite{BorkarMeyn} (see also section 3.2 of \cite{Bo08}) with a slight tweak which we return to after considering the asynchronous case, still under a.s.\ boundedness assumption. For the asynchronous case, by the arguments of \cite{asyn} (see also Chapter 7 of \cite{Bo08}), (\ref{odelim1}) gets modified to
    \begin{eqnarray}
    \dot{q}(t) &=& \Lambda(t)(F^{x}(q(t)) - q(t)) \nonumber \\
    &=& F^{x}_{\Lambda(t)}(q(t)) - q(t), \label{odelim2}
    \end{eqnarray}
where $\Lambda(t)$ is a diagonal matrix with  diagonal elements in $[0,1]$, and $F^{x}_{\Lambda}(q)$ for any such matrix $\Lambda$ is defined as $(I - \Lambda)q + \Lambda F^{x}(q)$. Furthermore, (\ref{eq:relsamplingfreq}) ensures that a.s., the diagonal elements of $\Lambda(t)$ remain uniformly bounded away from zero from below by some (possibly random) $\delta > 0$. The a.s.\ convergence again follows as in Theorem 3.5 of \cite{Abounadi2001learning} using the fact that the asymptotically stable equilibrium in the synchronous case remains the asymptotically stable equilibrium in this case as well even with the time dependence via $\Lambda(t)$. This is because $F^x_{\Lambda}$ have a common unique fixed point, viz., the unique fixed point $q^*$ of $F^x$, and a common Liapunov function $\|q - q^*\|_{\infty}$ applies. The a.s.\ boundedness argument argument of \cite{BorkarMeyn} (see also section 3.2 of \cite{Bo08}) applies exactly as in these references except that one has to account for the $t$-dependence through $\Lambda(t)$ and the $x$-dependence. The argument uses the scaled limit of the (\ref{odelim2}) given by
\begin{equation}
\dot{q}(t) = \hat{F}_{\Lambda(t)}^x(q(t)) - q(t), \label{odelim3}
\end{equation}
where $\hat{F}^x_{\Lambda}(q) := \lim_{\alpha\uparrow\infty}\frac{F^x_{\Lambda}(\alpha q)}{\alpha}$, which turns out to be of the same form as $F^x_{\Lambda}(q)$ except that the cost $\tilde{c}()$ in (\ref{asynlambda}) gets replaced by zero. Using arguments leading to Theorem 3.4 of \cite{Abounadi2001learning}, one verifies that the origin is the globally asymptotically stable equilibrium of (\ref{odelim3}) \textit{uniformly} in the $x$ variable\footnote{This $x$-dependence is the only additional feature (or point of difference) here as compared to \cite{Abounadi2001learning}.}, which takes values in a compact set because $\tilde{c}()$  is bounded, and $\Lambda$, as long as the diagonal elements thereof are uniformly bounded away from zero. A minor adaptation of the arguments of \cite{BorkarMeyn} (see also section 3.2 of \cite{Bo08}) yields the a.s.\ boundedness.

\item \textit{Step 2: Convergence of $\{\widetilde{Q}_n\}$:}

As before, since $\gamma_1(n) = o(\gamma_2(n))$, we invoke the two time scale arguments  of section 6.1, \cite{Bo08},  to treat both $x_n \approx x$ and $\epsilon_n \approx \epsilon$ as frozen. The proof of a.s.\ boundedness and a.s.\ convergence of $\{\widetilde{Q}_n\}$ is exactly the same as in `Step 1' above, i.e., as for Theorem 3.5 of \cite{Abounadi2001learning} with only one significant difference. The map $F^x$ above gets replaced by $\breve{F}^x$ where,
   \begin{displaymath}
    \breve{F}^{x}_{(s,a^1,a^2)}(q) :=  \tilde{c}(s, a^1, a^2; x) - f(q) - q((s,a^1),a^2)  + \sum_{(\tilde{s},\tilde{a}^1)}\tilde{p}((\tilde{s},\tilde{a}^1)|(s,a^1),a^2)\min_{b}\max_{b'}q((s,b),b').
    \end{displaymath}
The key property of $F^x$ used to prove Step 1 above is its non-expansivity with respect to the $\| \ \cdot \ \|_{\infty}$ norm. (See Lemma 3.1 of \cite{Abounadi2001learning}.) This property also holds for $\breve{F}^x$, so the arguments remain valid.

\item \textit{Step 3:  Convergence of $\{x_n\}$:}

From (\ref{eq:xn-learning-sg})-(\ref{epdecay}), it follows by arguments of Chapter 5, \cite{Bo08}, (see also \cite{BenHofSor1}, \cite{BenHofSor2}) that $\{x_n\}$ converges a.s.\ to an internally chain transitive invariant set of (\ref{eq:diffincleq1-sg}). In view of the proof of Theorem \ref{thm:bat-avggame}, this implies $x_n \to D$ a.s.

\item \textit{Step 4: Proof of optimality:}

This follows from a standard fact   about average cost MDPs recalled in the Appendix.

 \endproof

\end{itemize}

\begin{remark}
We can make a remark regarding convergence rate along the lines of the concluding remark of section 3. The additional complication here is the multiple time scales, the error analysis for which appears unavailable in literature. Nevertheless, it stands to reason that the convergence/sample complexity results be dictated by the slowest time scale which in this case is governed by the slowest stepsize $\gamma_1(n) = \frac{1}{n+1}$. Thus exactly the same observations as in the concluding remark of section 3 hold. There will be additional `tracking errors' from the fast time scale iterations, but these will remain within prescribed bounds with probability exceeding
$$1 - O\left(e^{-\frac{C'}{\sum_{m=n_0}^{\infty}\gamma_2(m)^2}}\right)$$
for $n \geq n_0$. Since $\gamma_1(n) = o(\gamma_2(n))$, this will be dominated by the error expression in (\ref{err}), so the latter continues to hold for the overall error.
\end{remark}

\section{Conclusion}
\label{sec:conclusion}
 We have presented a simple and computationally  tractable  strategy for approachability in  Stackelberg stochastic games.  We have also given a reinforcement learning based algorithm to learn the approachable strategy when the transition kernels are unknown. The motivation for this came from multi-objective optimization and decision making problems in a dynamically changing environment.

There are many interesting related questions that one can possibly address in the future.  Extension to   stochastic games with discounted reward is one problem but the solution is possibly very messy due to dependence on the initial state. Partial observations presents another challenge.

\ \\

\noindent \textbf{\large APPENDIX}

Consider  a controlled Markov chain $\{s_n\}$ with a finite state space $S$, compact metric action space $U$ with metric $d$, and running cost $k(i,u)$, with transition probabilities $p(j | i,u)$. Assume $k, p$ to be continuous in $u$. Also assume that Assumption 1 is true. The dynamic programming equation then is
\begin{displaymath}
V(i) = \min_u\left(k(i,u) - \kappa + \sum_jp(j | i,u)V(j)\right), \ i \in S.
\end{displaymath}
Then $\kappa$ is the optimal cost. Let $U^*(i)$ denote the set of minimizers on the right hand side, which will perforce be compact and nonempty by standard arguments. Suppose
$$d(a_n, U^*(s_n)) \to 0.$$
Then we have
\begin{eqnarray*}
V(s_n) - k(s_n, a_n) - \kappa + E[V(s_{n+1})|s_n, a_n] &\to& 0 \\
\Longrightarrow \ \lim_{n\uparrow\infty}\frac{1}{n}\sum_{m = 0}^{n-1}E[k(s_m, a_m)] - \kappa &=&  \lim_{n\uparrow\infty}\frac{1}{n}(E[V(s_{n})] - E[V(s_0)]) = 0.
\end{eqnarray*}
Hence $\{a_n\}$ is optimal. \\

\noindent \textbf{\large Acknowledgments.}
The authors are grateful to the anonymous reviewers for an outstanding job of refereeing, which has greatly improved the quality and readability of our paper.

 \bibliographystyle{IEEEtran}
\bibliography{References-Blackwell2}



\end{document}